\documentclass[11pt]{article}

\usepackage[preprint]{acl}

\usepackage{amsmath,amsfonts,bm}

\def\eqref#1{equation~\ref{#1}}

\def\1{\bm{1}}

\DeclareMathAlphabet{\mathsfit}{\encodingdefault}{\sfdefault}{m}{sl}
\SetMathAlphabet{\mathsfit}{bold}{\encodingdefault}{\sfdefault}{bx}{n}

\usepackage{times}
\usepackage{latexsym}
\usepackage[T1]{fontenc}
\usepackage[utf8]{inputenc}
\usepackage{microtype}
\usepackage{inconsolata}
\usepackage{graphicx}
\usepackage{hyperref}
\usepackage{url}
\usepackage{booktabs}
\usepackage[english]{babel}
\usepackage{amsfonts}
\usepackage{nicefrac}
\usepackage{multirow}
\usepackage[table]{xcolor}
\usepackage{float}
\usepackage{pifont}
\usepackage{amsmath}
\usepackage{amssymb}
\usepackage{makecell}
\usepackage{array}
\usepackage[most]{tcolorbox}
\usepackage{xcolor}
\newcolumntype{C}{>{\centering\arraybackslash}m{1.5em}}
\newcommand{\redstar}{\text{\textcolor{orange}{\ding{72}}}}
\newcommand{\best}[1]{\cellcolor{green!25}\textbf{#1}}

\usepackage{lineno}

\definecolor{darkblue}{rgb}{0, 0, 0.5}
\hypersetup{colorlinks=true, citecolor=darkblue, linkcolor=darkblue, urlcolor=darkblue}

\title{Structured Prompts Improve Evaluation of Language Models}

\author{\parbox{\textwidth}{\centering
  \makebox[0.95\textwidth]{\textbf{Asad Aali},\hfill \textbf{Muhammad Ahmed Mohsin},\hfill \textbf{Vasiliki Bikia},\hfill \textbf{Arnav Singhvi},}\\
  \makebox[0.95\textwidth]{\textbf{Richard Gaus},\hfill \textbf{Suhana Bedi},\hfill \textbf{Hejie Cui},\hfill \textbf{Miguel Fuentes},\hfill \textbf{Alyssa Unell},}\\
  \makebox[0.95\textwidth]{\textbf{Yifan Mai},\hfill \textbf{Jordan Cahoon},\hfill \textbf{Mike Pfeffer},\hfill \textbf{Roxana Daneshjou},\hfill \textbf{Sanmi Koyejo},}\\
  \makebox[0.95\textwidth]{\textbf{Emily Alsentzer},\hfill \textbf{Christopher Potts},\hfill \textbf{Nigam H. Shah},\hfill \textbf{Akshay S. Chaudhari}}\\[6pt]
  \textit{Stanford University}}}

\definecolor{baselineblue}{HTML}{1F4E79}
\definecolor{dspypurple}{HTML}{6A0DAD}
\definecolor{contextgreen}{HTML}{1B5E20}
\definecolor{miprorange}{HTML}{E65100}

\newcommand{\dspy}[1]{\textcolor{dspypurple}{#1}}
\newcommand{\context}[1]{\textcolor{contextgreen}{#1}}
\newcommand{\mipro}[1]{\textcolor{miprorange}{#1}}

\begin{document}
\maketitle

\begin{abstract}
  As language models (LMs) are increasingly adopted across domains, high-quality benchmarking frameworks are essential for guiding deployment decisions. However, frameworks such as Holistic Evaluation of Language Models (HELM) typically evaluate models under a single static prompt configuration, even though model behavior depends strongly on prompt choice. As a result, reported scores can reflect prompt choice as much as model capability. Declarative prompting frameworks, such as DSPy, offer a scalable way to operationalize this evaluation protocol by evaluating models under structured prompting strategies rather than static prompts. Here, we conduct a reproducible \textit{DSPy+HELM} benchmarking study and quantify how prompt choice impacts benchmark outcomes. Using five prompting methods, we evaluate four frontier and two open-source LMs across seven benchmarks against HELM baseline scores and find that prompt choice can materially impact leaderboards. Structured prompting improves performance (by 6\% on average), alters comparisons (leaderboard rankings shift on $5/7$ benchmarks); \textit{chain-of-thought} provides the strongest cost-performance tradeoff while more advanced optimizers offer smaller additional gains. Beyond the known intuition that prompts affect outputs, our benchmarking study integrates structured prompting into an established evaluation framework and quantifies how prompt choice alone impacts leaderboards.\footnote{We open-source (i) \href{https://github.com/stanford-crfm/helm/pull/3893}{\textit{DSPy+HELM} Evaluation} and (ii) \href{https://github.com/StanfordMIMI/dspy-helm}{Prompt Optimization Pipeline}.}.
\end{abstract}

\section{Introduction}
Language models (LMs) have advanced in text generation, spurring deployment across domains~\citep{thirunavukarasu2023large, van2024adapted, seo2024evaluation}. Yet, integrating LMs into downstream workflows is challenging as LMs frequently commit errors~\citep{aali2025medval}. Even state-of-the-art frontier LMs exhibit non-trivial hallucination rates~\citep{wang2024prompt, sivarajkumar2024empirical, bang2025hallulens, tamber2025benchmarking}. Such concerns are compounded by LMs’ sensitivity to prompt design~\citep{razavi2025benchmarking}.

While benchmarking frameworks such as Holistic Evaluation of Language Models (HELM)~\citep{liang2022holistic, bedi2025medhelm} enable evaluation of LMs across tasks, public leaderboards typically evaluate multiple LMs under a single static prompt configuration. Because LMs respond differently to different prompts, a static prompt measures performance under one particular invocation rather than fully characterizing model capability. This can distort both absolute scores and relative rankings. Hence, broader adoption necessitates protocols that move beyond a static prompt configuration.

Prompt engineering has emerged as a valuable strategy for improving model performance~\citep{nori2024medprompt, maharjan2024openmedlm}, combining few-shot selection, chain-of-thought (CoT)~\citep{wei2022chainofthought}, and ensembling. However, these methods rely on hand-engineered prompts and demand iterative experimentation, making them labor-intensive and brittle~\citep{wang2025perspective}. Consequently, researchers have explored automatic prompt optimization (APO)~\citep{li2025surveyautomaticpromptengineering}, which treats prompt design as an optimization problem. DSPy~\citep{khattab2023dspy} is a widely used framework that represents prompts as modular, parameterized components with an intuitive structure that allows moving from zero-shot prompts to more adaptive prompt optimizers~\citep{opsahl2024optimizing} all within a single unified system.

\begin{figure}[t]
\centering
\includegraphics[width=0.48\textwidth]{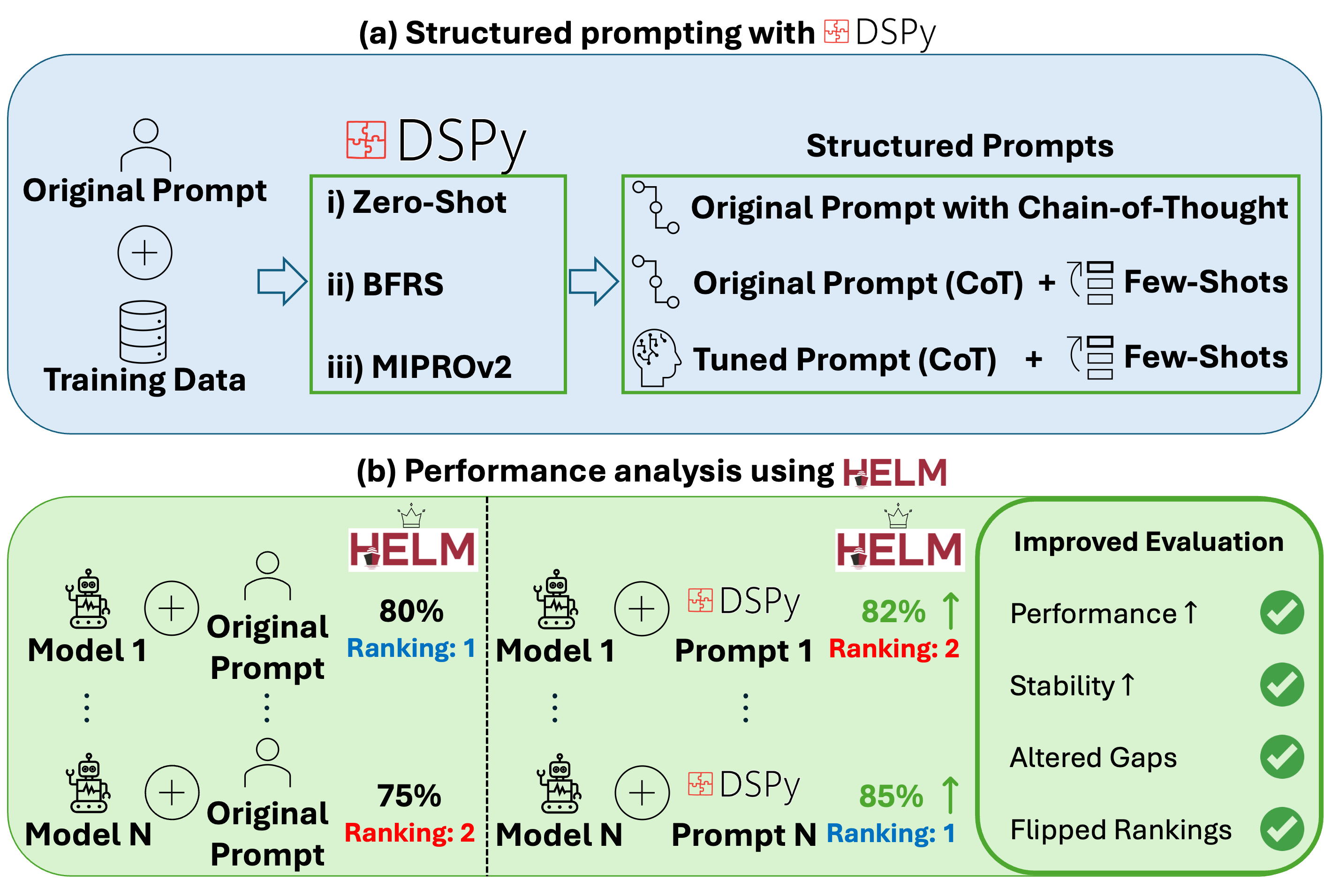}
\footnotesize
\caption{Pipeline overview. (a) DSPy takes HELM's baseline prompt and produces structured prompt variants. (b) HELM evaluates models under each prompt variant. With structured prompting, evaluation improves and benchmark conclusions change materially.}
\label{fig:method}
\end{figure}

Prior work shows that instruction paraphrases can alter scores and rankings~\citep{mizrahi2024state}. In this work, we use DSPy as an instantiation of structured prompting and integrate it with HELM (Figure~\ref{fig:method}) to examine how structured prompting affects benchmark outcomes, presenting:
\begin{enumerate}
    \item A reproducible \textit{DSPy+HELM} evaluation protocol for systematically studying how prompt choice affects benchmark conclusions.
    \item An evaluation of five prompting methods (HELM Baseline, Zero-Shot Predict, Zero-Shot CoT, BFRS, MIPROv2) across four frontier LMs, two open-source LMs and seven benchmarks spanning diverse domains.
    \item Empirical evidence that prompt choice can materially impact leaderboards. Structured prompting improves performance (by 6\% on average), alters comparisons (leaderboard rankings shift on $5/7$ benchmarks); \textit{chain-of-thought} provides the strongest cost-performance tradeoff while more advanced optimizers offer smaller additional gains.
\end{enumerate}

\section{Methodology}

DSPy~\citep{khattab2023dspy} is a framework for composing modular LM pipelines. Formally, let $\Phi$ denote a LM program with $m$ modules. Each module $i$ has a prompt template $p_i$ containing a set of variables for the instruction and $K$ demonstration examples. Let $V$ be the set of all such prompt variables across $\Phi$, and let $V \to S$ denote an assignment of each variable $v \in V$ to a concrete string $s \in S$. We write $\Phi_{V \to S}$ to denote running program $\Phi$ under a particular prompt. Given a dataset $D = {(x, y)}$ of inputs $x$ with ground-truth $y$ and an evaluation metric $\mu$ that compares the program’s output $\Phi(x)$ against $y$, the optimization maximizes $\mu$ over all instructions and demonstrations.

\begin{figure}[t]
\centering
\includegraphics[width=0.48\textwidth]{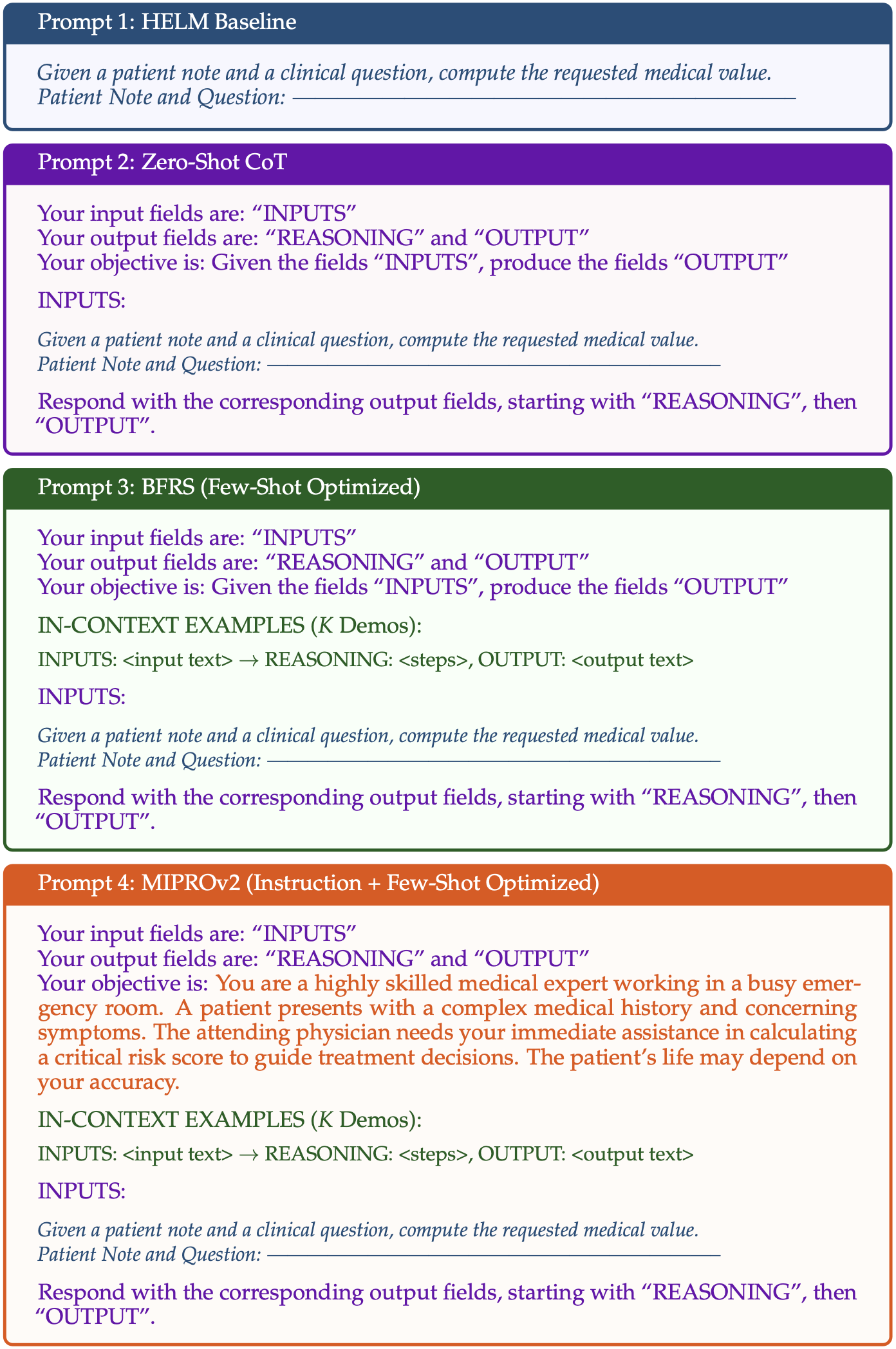}
\footnotesize
\caption{Structured prompting methods evaluated in our study (\dspy{Zero-Shot CoT}, \context{BFRS}, \mipro{MIPROv2}). Each box corresponds to a method showing how prompts differ.}
\label{fig:prompts}
\end{figure}

\begin{table*}[t]
\centering
\small
\setlength{\tabcolsep}{18.5pt}
\renewcommand{\arraystretch}{1.15}
\begin{tabular}{lllc}
\toprule
\textbf{Benchmark} &
\textbf{Input $\rightarrow$ Output} &
\textbf{Task} &
\textbf{Samples} \\
\midrule
MMLU-Pro &
Reasoning Question $\rightarrow$ Answer & Multi-Task Reasoning & 1,000 \\
GPQA & Graduate Question $\rightarrow$ Answer & Graduate-Level QA &
446 \\
GSM8K &
Math Problem $\rightarrow$ Solution & Numeric Problem-Solving & 1,000 \\
MedCalc-Bench &
Patient Note $\rightarrow$ Computed Value & Computational Reasoning & 1,000 \\
Medec & Medical Narrative $\rightarrow$ Errors & Error Classification &
597 \\
HeadQA &
Medical Question $\rightarrow$ Answer &
USMLE-Style QA &
1,000 \\
MedBullets &
Medical Question $\rightarrow$ Answer &
USMLE-Style QA &
308 \\
\bottomrule
\end{tabular}
\caption{HELM benchmarks evaluated in our study. Columns summarize the number of test samples and tasks spanning reasoning, knowledge QA, problem-solving, and error-classification tasks across diverse domains.}
\label{tab:dataset-descriptions}
\end{table*}

\begin{table*}[t]
\centering
\small
\setlength{\tabcolsep}{5.1pt}
\renewcommand{\arraystretch}{1.15}
\begin{tabular}{llcccc}
\toprule
\textbf{Model} & \textbf{API Identifier} & \textbf{Release} & \textbf{Context} & \textbf{Reasoning} & \textbf{Open-Source}\\
\midrule
Claude 3.7 Sonnet & \texttt{anthropic/claude-3-7-sonnet-20250219} & 02/19/2025 & 200k & \ding{55} & \ding{55} \\
Gemini 2.0 Flash & \texttt{google/gemini-2.0-flash-001} & 02/01/2025 & 1000k & \ding{55} & \ding{55}\\
GPT 4o & \texttt{openai/gpt-4o-2024-05-13} & 05/13/2024 & 128k & \ding{55} & \ding{55}\\
o3 Mini & \texttt{openai/o3-mini-2025-01-31} & 01/31/2025 & 200k & \ding{51} & \ding{55}\\
Llama 3.3 70B & \texttt{meta-llama/Llama-3.3-70B-Instruct} & 12/06/2024 & 128k & \ding{55} & \ding{51}\\
Qwen3 4B & \texttt{Qwen/Qwen3-4B-Instruct-2507} & 08/06/2025 & 256k & \ding{55} & \ding{51}\\
\bottomrule
\end{tabular}
\caption{Evaluated language models with their API identifier, release, context, reasoning mode, and availability.}
\label{tab:models}
\end{table*}

\subsection{Prompting Methods}

We evaluate these prompting methods (Figure~\ref{fig:prompts}):

\textbf{1. HELM Baseline. }HELM supports multiple prompting configurations; we adopt the commonly reported zero-shot configuration without CoT. We treat this baseline as one configuration among others, with the objective of measuring prompt sensitivity rather than outperforming this baseline.

\textbf{2. Zero-Shot Predict. }Zero-Shot Predict is an unoptimized non-adaptive baseline, instantiated with the \texttt{dspy.Predict} module. Each module uses HELM's baseline instruction, but under DSPy's interface without demonstrations (i.e.\ $K=0$).

\textbf{3. Zero-Shot CoT. } Zero-Shot CoT utilizes the same prompting structure as Zero-Shot Predict, but instantiates the \texttt{dspy.ChainOfThought} module, which elicits step-by-step rationales, instructing the LM to generate a reasoning trace.

\textbf{4. BFRS. }Bootstrap Few-Shot with Random Search (BFRS) leverages the idea of bootstrapping to select the best demonstrations in two phases: (i) Bootstrapping demonstrations: the LM program $\Phi$ is run on a subset of inputs to gather traces. Whenever the output of $\Phi(x)$ for an example $x$ achieves a sufficiently high score (on metric $\mu$), the input-output pair is taken as a candidate demonstration. (ii) Random few-shot search: Given demonstration pools, BFRS randomly samples sets of $K$ demonstrations per module, inserts them into the module, and evaluates the program on a validation split. After $N$ combinations, the highest scoring program is returned (with hyperparameters $K$ and $N$).

\textbf{5. MIPROv2. }MIPROv2 is an optimizer that selects instructions and $K$ few-shot demonstrations via: (i) bootstrapping demos, (ii) grounded instruction proposals from a proposer LM conditioned on dataset summaries, program structure, demos, and trial history, and (iii) Bayesian search over instruction-demo pairs. It treats each configuration $\mathbf{v}$ as hyperparameters, learns $p(y \mid \mathbf{v})$ from outcomes, and steers toward high-scoring regions. The best prompt configuration is returned (with hyperparameters instruction text, demo-set, and $K$).

\subsection{Benchmarks}
We choose benchmarks (Table~\ref{tab:dataset-descriptions}) based on (i) availability, (ii) diversity (reasoning, QA, problem-solving), and (iii) domains (general/medical).

\textbf{MMLU-Pro. }MMLU-Pro~\citep{wang2024mmlu} is an enhanced version of MMLU that focuses on challenging reasoning questions, providing a discriminative measure of reasoning.

\textbf{GPQA. }GPQA~\citep{rein2024gpqa} is a graduate-level multiple-choice benchmark covering biology, physics, and chemistry to test advanced reasoning.

\textbf{GSM8K. }GSM8K~\citep{cobbe2021gsm8k} contains grade school math word problems. The task requires computing a final numeric answer.

\textbf{MedCalc-Bench. }MedCalc-Bench~\citep{khandekar2024medcalc} is a medical calculation benchmark, where the input is a patient note and a question asking for a categorical value.

\textbf{Medec. }Medec~\citep{abacha2024medec} is an error detection benchmark, where inputs contain clinical narratives, and the task is to identify/correct errors.

\textbf{HeadQA. } HeadQA~\citep{vilares2019head} is a collection of biomedical multiple-choice questions, where questions cover knowledge resembling board exams.

\textbf{MedBullets. }MedBullets~\citep{medbullets} is a benchmark of USMLE-style questions with multiple-choice answers designed to reflect the difficulty of medical licensing exams.

\begin{table*}[t]
\centering
\scriptsize
\setlength{\tabcolsep}{12.5pt}
\renewcommand{\arraystretch}{1.15}
\begin{tabular}{l@{\hspace{12pt}}cccccc}
\toprule
\textbf{Prompting Method} &
\shortstack{\textbf{Claude 3.7}\\\textbf{Sonnet}} &
\shortstack{\textbf{Gemini 2.0}\\\textbf{Flash}} &
\shortstack{\textbf{GPT}\\\textbf{4o}} &
\shortstack{\textbf{o3}\\\textbf{Mini}} &
\shortstack{\textbf{Llama 3.3}\\\textbf{70B}} &
\shortstack{\textbf{Qwen3}\\\textbf{4B}} \\
\midrule
HELM Baseline
& 64.8\% $\pm$ 3.3 & 61.4\% $\pm$ 3.4 & 61.0\% $\pm$ 3.3 & 70.9\% $\pm$ 3.3 & 57.2\% $\pm$ 3.4 & 47.8\% $\pm$ 3.3 \\
\cmidrule{1-7}
Zero-Shot Predict & 65.1\% $\pm$ 3.4 & 61.7\% $\pm$ 3.4 & 59.7\% $\pm$ 3.3 & \best{73.2\% $\pm$ 3.2} & 59.2\% $\pm$ 3.4 & 49.3\% $\pm$ 3.4 \\
\noalign{\vskip 2pt}
Zero-Shot CoT \redstar & 69.4\% $\pm$ 3.2 & \best{66.2\% $\pm$ 3.4} & 65.7\% $\pm$ 3.3 & 72.7\% $\pm$ 3.2 & 64.0\% $\pm$ 3.4 & 59.0\% $\pm$ 3.4 \\
BFRS \redstar    & 69.3\% $\pm$ 3.3 & \best{66.2\% $\pm$ 3.3} & \best{65.9\% $\pm$ 3.2} & 73.1\% $\pm$ 3.2 & \best{64.2\% $\pm$ 3.3} & 60.3\% $\pm$ 3.4\\
MIPROv2 \redstar  & \best{69.8\% $\pm$ 3.3} & \best{66.2\% $\pm$ 3.3} & 65.3\% $\pm$ 3.3 & 73.1\% $\pm$ 3.1 & 62.1\% $\pm$ 3.4 & \best{60.4\% $\pm$ 3.4} \\
\noalign{\vskip 4pt}
\midrule
\midrule
\textbf{Highest $-$ Baseline \redstar}
& {+5.0\%} & {+4.8\%} & {+4.9\%} & {+2.3\%} & {+7.0\%} & {+12.6\%} \\
\bottomrule
\end{tabular}
\caption{HELM leaderboard (macro-averaged) across six language models and five prompting methods. \textbf{Green} marks the highest performance among evaluated prompting methods. Entries are reported as mean $\pm$ 95\% bootstrap confidence interval. \redstar~indicates a statistically significant pooled improvement over the HELM baseline.}
\label{tab:average_benchmarks}
\end{table*}

\subsection{Experimental Setup}

We evaluate six LMs (Table~\ref{tab:models}) and initialize each DSPy program with HELM’s baseline instruction. For BFRS and MIPROv2, we follow DSPy’s data separation: demonstrations are bootstrapped \emph{exclusively} from training data and candidate prompts are evaluated on a \emph{disjoint} validation split from the training data; neither optimizer ever sees the HELM leaderboard test set. We create fixed train/val partitions (GPQA 358/90, GSM8K 6725/100, HeadQA 2391/100, MedBullets 246/62, MedCalc-Bench 9047/100, MEDEC 1970/100, MMLU-Pro 35/35) where validation is capped at 100 for larger datasets. Our pools satisfy DSPy's intended low-data usecase for optimizers: BFRS $\geq50$ examples and MIPROv2 $\geq200$ (except MMLU-Pro). Demonstrations are capped at $K\leq3$ per module. BFRS uses 16 optimization trials and MIPROv2 uses 13. All final scoring is performed via HELM, so outputs are judged identically. All results reflect single, deterministic runs (temperature = 0), matching HELM's setup. For baselines, we report HELM's public leaderboard scores when the setup matches ours: (i) identical LM API version, (ii) zero-shot prompting, and (iii) no CoT. For benchmarks where the leaderboard setup does not match, we reproduce them with single, deterministic runs. To assess whether gains reflect systematic improvements rather than noise, we apply an exact two-sided McNemar test to paired correct/incorrect outcomes, with Benjamini-Hochberg FDR correction for multiple comparisons. Under a pooled analysis aggregating all benchmark instances across models and benchmarks, each structured prompting method (Zero-Shot CoT, BFRS, and MIPROv2) shows a statistically significant improvement over baseline.

\section{Results and Discussion}

\subsection{Impact of Structured Prompting}
\label{sec:empirical_results}

\begin{table*}[t]
\centering
\scriptsize
\setlength{\tabcolsep}{8.8pt}
\renewcommand{\arraystretch}{1.15}
\begin{tabular}{ll@{\hspace{12pt}}cccccc}
\toprule
\textbf{Benchmark} & \textbf{Prompting Method} &
\shortstack{\textbf{Claude 3.7}\\\textbf{Sonnet}} &
\shortstack{\textbf{Gemini 2.0}\\\textbf{Flash}} &
\shortstack{\textbf{GPT}\\\textbf{4o}} &
\shortstack{\textbf{o3}\\\textbf{Mini}} &
\shortstack{\textbf{Llama 3.3}\\\textbf{70B}} &
\shortstack{\textbf{Qwen3}\\\textbf{4B}} \\
\midrule
\multirow{5}{*}{\shortstack{MMLU-Pro}} & HELM Baseline
& 76.3\% $\pm$ 2.7 & 66.1\% $\pm$ 3.0 & 62.2\% $\pm$ 3.0 & 77.1\% $\pm$ 3.1 & 64.7\% $\pm$ 3.0 & 44.9\% $\pm$ 3.1 \\
\cmidrule{2-8}
& Zero-Shot Predict & 77.7\% $\pm$ 2.7 & 70.3\% $\pm$ 2.8 & 60.7\% $\pm$ 3.0 & \best{78.4\% $\pm$ 3.1} & 62.7\% $\pm$ 3.0 & 41.7\% $\pm$ 3.0 \\
\noalign{\vskip 2pt}
& Zero-Shot CoT & 79.7\% $\pm$ 2.5 & 75.3\% $\pm$ 2.8 & 67.6\% $\pm$ 3.0 & 76.2\% $\pm$ 3.1 & \best{68.5\% $\pm$ 2.9} & 66.2\% $\pm$ 3.0 \\
& BFRS    & 80.1\% $\pm$ 2.5 & \best{75.4\% $\pm$ 2.7} & \best{71.1\% $\pm$ 2.8} & 76.5\% $\pm$ 3.1 & \best{68.5\% $\pm$ 2.9} & \best{68.6\% $\pm$ 2.9} \\
& MIPROv2  & \best{80.6\% $\pm$ 2.4} & 75.3\% $\pm$ 2.6 & 68.7\% $\pm$ 2.9 & 76.1\% $\pm$ 3.1 & 56.5\% $\pm$ 3.1 & \best{68.6\% $\pm$ 2.9} \\
\midrule
\multirow{5}{*}{GPQA} & HELM Baseline
& 57.0\% $\pm$ 4.7 & 53.4\% $\pm$ 4.7 & 45.5\% $\pm$ 4.7 & 57.6\% $\pm$ 4.7 & 40.4\% $\pm$ 4.5 & 34.3\% $\pm$ 4.3 \\
\cmidrule{2-8}
& Zero-Shot Predict & 62.1\% $\pm$ 4.7 & 54.5\% $\pm$ 4.7 & 41.7\% $\pm$ 4.5 & 66.6\% $\pm$ 4.5 & 55.8\% $\pm$ 4.7 & 35.2\% $\pm$ 4.5 \\
\noalign{\vskip 2pt}
& Zero-Shot CoT  & 61.4\% $\pm$ 4.5 & 59.2\% $\pm$ 4.7 & \best{52.5\% $\pm$ 4.7} & 66.4\% $\pm$ 4.7 & 55.8\% $\pm$ 4.7 & 50.0\% $\pm$ 4.7 \\
& BFRS     & \best{64.1\% $\pm$ 4.5} & \best{61.0\% $\pm$ 4.5} & 49.3\% $\pm$ 4.7 & 65.5\% $\pm$ 4.5 & \best{56.3\% $\pm$ 4.7} & 47.8\% $\pm$ 4.7 \\
& MIPROv2  & 61.9\% $\pm$ 4.5 & 59.0\% $\pm$ 4.7 & 47.8\% $\pm$ 4.7 & \best{68.4\% $\pm$ 4.3} & 52.7\% $\pm$ 4.7 & \best{52.2\% $\pm$ 4.7} \\
\midrule
\multirow{5}{*}{GSM8K} & HELM Baseline
& 80.5\% $\pm$ 2.5 & 84.0\% $\pm$ 2.3 & 81.1\% $\pm$ 2.4 & 88.6\% $\pm$ 2.0 & 85.1\% $\pm$ 2.2 & 80.2\% $\pm$ 2.5 \\
\cmidrule{2-8}
& Zero-Shot Predict & 83.0\% $\pm$ 2.3 & 77.3\% $\pm$ 2.7 & 84.6\% $\pm$ 2.3 & \best{93.6\% $\pm$ 1.6} & 86.8\% $\pm$ 2.1 & 84.1\% $\pm$ 2.3 \\
\noalign{\vskip 2pt}
& Zero-Shot CoT  & 83.3\% $\pm$ 2.3 & 83.1\% $\pm$ 2.4 & \best{90.7\% $\pm$ 1.9} & 92.6\% $\pm$ 1.6 & 89.0\% $\pm$ 2.0 & 88.7\% $\pm$ 2.0 \\
& BFRS      & 83.2\% $\pm$ 2.3 & \best{84.2\% $\pm$ 2.3} & 90.4\% $\pm$ 1.8 & 93.0\% $\pm$ 1.6 & 90.0\% $\pm$ 1.9 & \best{91.9\% $\pm$ 1.8} \\
& MIPROv2   & \best{84.0\% $\pm$ 2.3} & 83.5\% $\pm$ 2.3 & 89.8\% $\pm$ 1.9 & 93.4\% $\pm$ 1.6 & \best{90.8\% $\pm$ 1.8} & 90.7\% $\pm$ 1.8 \\
\bottomrule
\end{tabular}
\caption{HELM leaderboard (general) across six language models and five prompting methods. \textbf{Green} marks the highest performance among evaluated prompting methods. Entries are reported as mean $\pm$ 95\% bootstrap CI.}
\label{tab:helm_benchmarks}
\end{table*}

\textbf{Improved performance.} Structured prompting improves over HELM baseline (Table \ref{tab:average_benchmarks}). Under the best-performing prompt, average accuracy increases by 6\% across models. Non-reasoning models benefit most, while \textit{o3 Mini} sees smaller gains.

\textbf{Shifted leaderboard rankings.} Heterogeneous gains shift rankings on $5/7$ benchmarks under the best-performing prompt. Notably, on MMLU-Pro (Table~\ref{tab:helm_benchmarks}), baseline \textit{o3 Mini > Claude 3.7 Sonnet} (77.1\% vs.\ 76.3\%) changes to \textit{Claude 3.7 Sonnet > o3 Mini} (80.6\% vs.\ 78.4\%). On GSM8K, \textit{GPT 4o} overtakes \textit{Gemini 2.0 Flash}, shifting from (81.1\% vs.\ 84.0\%) to (90.7\% vs.\ 84.2\%). On MedCalc-Bench (Table~\ref{tab:medhelm_benchmarks}), baseline \textit{o3 Mini > Claude 3.7 Sonnet} (34.0\% vs.\ 21.0\%) becomes \textit{Claude 3.7 Sonnet > o3 Mini} (35.3\% vs.\ 34.7\%).

\begin{figure}[t]
  \centering
  \includegraphics[width=0.48\textwidth]{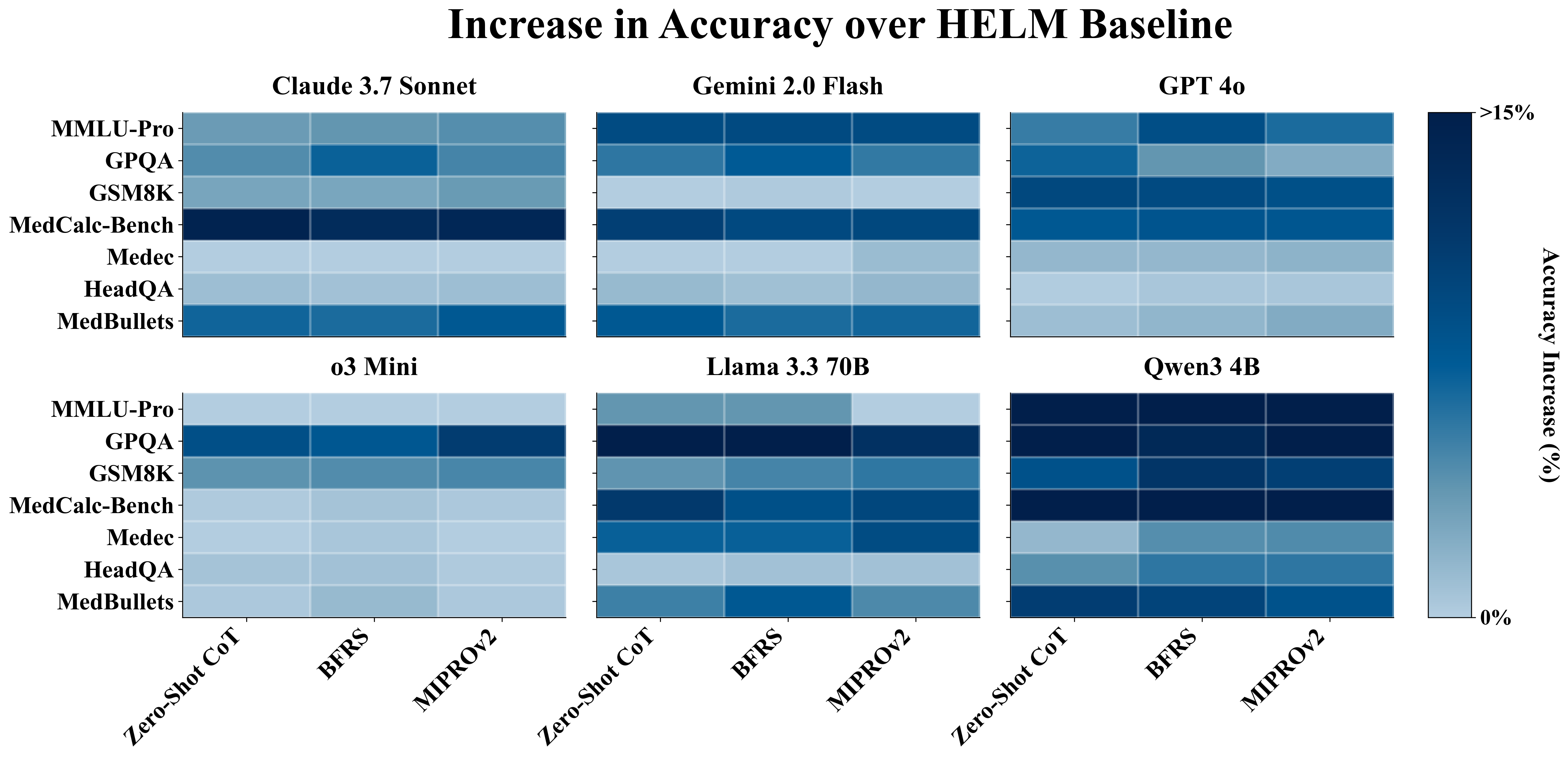}
  \footnotesize
  \caption{Heat map showing increase in accuracy of each prompting method over HELM's baseline (light=small, dark=large). Across six models, x-axis lists prompting methods, y-axis lists benchmarks. All structured prompting methods exhibit similar improvements, \textit{o3 Mini} remains relatively insensitive, and smaller open-source models achieve largest gains.}
  \label{fig:performance_improvements}
\end{figure}

\textbf{Altered performance gaps.} When evaluated under the best-performing prompt, models can either narrow or widen their relative performance gaps, providing a better view of capability differences. Averaging across benchmarks, the gap between the top two models (\textit{o3 Mini} and \textit{Claude 3.7 Sonnet}) shrinks from 6\% at baseline (70.9\% vs.\ 64.8\%) to 3\% (73.2\% vs.\ 69.8\%). However, this trend is not uniform: on GPQA, the gap widens substantially, from 0.6\% at baseline (57.6\% vs.\ 57.0\%) to 4.3\% under the best performance (68.4\% vs.\ 64.1\%).

\textbf{Benchmark-dependent sensitivity.} Reasoning tasks such as MMLU-Pro, GPQA, GSM8K, MedCalc-Bench, and MedBullets, show larger gains (Figure~\ref{fig:performance_improvements}). In contrast, HeadQA and Medec exhibit smaller gains. We hypothesize that HeadQA is bottlenecked by saturated baselines, while Medec likely reflects knowledge gaps.

\textbf{Model-dependent sensitivity.} Prompt sensitivity varies by model regime: \textit{o3 Mini} gains less beyond CoT, whereas smaller open models gain more from structured prompting. Notably, on MMLU-Pro, GSM8K, and MedCalc-Bench, \textit{Qwen3 4B} slightly exceeds \textit{Llama 3.3 70B} under the best-performing prompt despite being $18\times$ smaller.

\textbf{Ranking stability.} To assess how prompt choice shifts relative rankings, we compute mean ranks (1 = best, 6 = worst) across all benchmarks. Under the best-performing prompt, \textit{o3 Mini} remains the top model but becomes less dominant (1.29 $\to$ 1.57), \textit{Claude 3.7 Sonnet} improves modestly (2.43 $\to$ 2.29), and \textit{Qwen3 4B} also improves materially (6.00 $\to$ 5.14). In contrast, \textit{GPT 4o} (3.43 $\to$ 3.57), \textit{Gemini 2.0 Flash} (3.43 $\to$ 3.86), and \textit{Llama 3.3 70B} (4.43 $\to$ 4.57) decline slightly. Rank standard deviation ($\sigma$) shows a similar pattern: \textit{o3 Mini} remains roughly unchanged (0.76 $\to$ 0.79), while \textit{Claude 3.7 Sonnet} (1.27 $\to$ 1.80), \textit{GPT 4o} (1.13 $\to$ 1.27), \textit{Gemini 2.0 Flash} (0.53 $\to$ 0.69), \textit{Llama 3.3 70B} (1.13 $\to$ 1.27), and \textit{Qwen3 4B} (0.00 $\to$ 1.46) become more variable. We additionally perform instance-level bootstrap resampling, recomputing ranks under both baseline and structured-prompt settings. While some 95\% bootstrap intervals overlap, the direction of the rank shifts remains stable.

\textbf{CoT reduces sensitivity to prompt design.} We study the impact of each prompting method on the leaderboard by averaging results across LMs and benchmarks. Moving from HELM’s baseline to Zero-Shot Predict yields minimal improvement (60.5\% $\to$ 61.4\%). In contrast, introducing CoT reasoning and moving from Zero-Shot Predict to Zero-Shot CoT results in statistically significant gains (61.4\% $\to$ 66.2\%). Once CoT is introduced, moving to more sophisticated optimizers (BFRS and MIPROv2) does not lead to additional improvements (66.2\% $\to$ 66.3\%). We report this as an empirical finding rather than a theoretical claim.

\begin{table*}[t]
\centering
\scriptsize
\setlength{\tabcolsep}{8pt}
\renewcommand{\arraystretch}{1.15}
\begin{tabular}{ll@{\hspace{12pt}}cccccc}
\toprule
\textbf{Benchmark} & \textbf{Prompting Method} &
\shortstack{\textbf{Claude 3.7}\\\textbf{Sonnet}} &
\shortstack{\textbf{Gemini 2.0}\\\textbf{Flash}} &
\shortstack{\textbf{GPT}\\\textbf{4o}} &
\shortstack{\textbf{o3}\\\textbf{Mini}} &
\shortstack{\textbf{Llama 3.3}\\\textbf{70B}} &
\shortstack{\textbf{Qwen3}\\\textbf{4B}} \\
\midrule
\multirow{5}{*}{\shortstack{MedCalc-Bench}} & HELM Baseline
  & 21.0\% $\pm$ 2.5 & 15.8\% $\pm$ 2.3 & 18.8\% $\pm$ 2.4 & 34.0\% $\pm$ 2.9 & 11.3\% $\pm$ 2.0 & 4.7\% $\pm$ 1.4 \\
\cmidrule{2-8}
& Zero-Shot Predict & 20.6\% $\pm$ 2.5 & 17.0\% $\pm$ 2.4 & 15.7\% $\pm$ 2.3 & 33.4\% $\pm$ 3.0 & 9.9\% $\pm$ 1.9 & 11.5\% $\pm$ 2.0 \\
\noalign{\vskip 2pt}
& Zero-Shot CoT & \best{35.3\% $\pm$ 3.0} & \best{26.3\% $\pm$ 2.7} & 26.6\% $\pm$ 2.8 & 34.2\% $\pm$ 3.0 & \best{22.5\% $\pm$ 2.6} & 20.8\% $\pm$ 2.5 \\
& BFRS   & 34.1\% $\pm$ 3.0 & 25.2\% $\pm$ 2.7 & \best{27.0\% $\pm$ 2.8} & \best{34.7\% $\pm$ 3.0} & 20.0\% $\pm$ 2.5 & \best{22.7\% $\pm$ 2.6} \\
& MIPROv2 & 34.7\% $\pm$ 3.0 & 25.4\% $\pm$ 2.8 & 26.8\% $\pm$ 2.8 & 34.3\% $\pm$ 2.9 & 21.0\% $\pm$ 2.5 & 21.5\% $\pm$ 2.5 \\
\midrule
\multirow{5}{*}{Medec} & HELM Baseline
& \best{62.8\% $\pm$ 3.9} & 59.6\% $\pm$ 4.0 & 58.0\% $\pm$ 3.9 & 68.7\% $\pm$ 3.7 & 52.9\% $\pm$ 4.0 & 52.1\% $\pm$ 4.0 \\
\cmidrule{2-8}
& Zero-Shot Predict & 58.3\% $\pm$ 4.0 & 59.3\% $\pm$ 4.0 & 57.3\% $\pm$ 4.0 & 68.3\% $\pm$ 3.7 & 53.6\% $\pm$ 4.0 & 52.3\% $\pm$ 4.0 \\
\noalign{\vskip 2pt}
& Zero-Shot CoT & 61.8\% $\pm$ 3.9 & 59.5\% $\pm$ 3.9 & 59.5\% $\pm$ 4.0 & 68.2\% $\pm$ 3.7 & 60.1\% $\pm$ 4.0 & 53.6\% $\pm$ 4.0 \\
& BFRS   & 60.5\% $\pm$ 4.0 & 59.1\% $\pm$ 3.9 & 59.5\% $\pm$ 3.9 & \best{69.2\% $\pm$ 3.7} & 60.1\% $\pm$ 4.0 & 56.4\% $\pm$ 4.0 \\
& MIPROv2 & 62.5\% $\pm$ 4.0 & \best{60.8\% $\pm$ 3.9} & \best{59.8\% $\pm$ 4.0} & 68.3\% $\pm$ 3.7 & \best{62.0\% $\pm$ 4.0} & \best{56.6\% $\pm$ 4.0} \\
\midrule
\multirow{5}{*}{HeadQA} & HELM Baseline
& 91.2\% $\pm$ 1.8 & 88.0\% $\pm$ 2.1 & 90.6\% $\pm$ 1.9 & 89.3\% $\pm$ 2.0 & 85.4\% $\pm$ 2.3 & 76.7\% $\pm$ 2.6 \\
\cmidrule{2-8}
& Zero-Shot Predict & 88.7\% $\pm$ 2.0 & 88.5\% $\pm$ 2.0 & 86.4\% $\pm$ 2.1 & \best{90.9\% $\pm$ 1.8} & 81.7\% $\pm$ 2.4 & 76.9\% $\pm$ 2.7 \\
\noalign{\vskip 2pt}
& Zero-Shot CoT & \best{92.2\% $\pm$ 1.7} & 89.3\% $\pm$ 1.9 & 90.7\% $\pm$ 1.8 & 90.0\% $\pm$ 1.9 & 85.9\% $\pm$ 2.2 & 80.9\% $\pm$ 2.4 \\
& BFRS     & 92.0\% $\pm$ 1.7 & 88.9\% $\pm$ 2.0 & \best{91.1\% $\pm$ 1.8} & 90.1\% $\pm$ 1.9 & \best{86.2\% $\pm$ 2.2} & \best{82.5\% $\pm$ 2.4} \\
& MIPROv2   & \best{92.2\% $\pm$ 1.7} & \best{89.5\% $\pm$ 1.9} & \best{91.1\% $\pm$ 1.8} & 89.5\% $\pm$ 1.9 & \best{86.2\% $\pm$ 2.2} & \best{82.5\% $\pm$ 2.4} \\
\midrule
\multirow{5}{*}{MedBullets} & HELM Baseline
& 64.9\% $\pm$ 5.2 & 63.0\% $\pm$ 5.5 & 71.1\% $\pm$ 4.9 & 81.2\% $\pm$ 4.5 & 60.7\% $\pm$ 5.5 & 41.9\% $\pm$ 5.5 \\
\cmidrule{2-8}
& Zero-Shot Predict & 65.3\% $\pm$ 5.5 & 64.9\% $\pm$ 5.5 & 71.4\% $\pm$ 5.2 & 81.5\% $\pm$ 4.5 & 63.6\% $\pm$ 5.5 & 43.5\% $\pm$ 5.5 \\
\noalign{\vskip 2pt}
& Zero-Shot CoT   & 71.8\% $\pm$ 4.9 & \best{70.8\% $\pm$ 5.2} & 72.1\% $\pm$ 5.2 & 81.5\% $\pm$ 4.2 & 65.9\% $\pm$ 5.5 & \best{52.6\% $\pm$ 5.5} \\
& BFRS  & 71.4\% $\pm$ 5.2 & 69.5\% $\pm$ 5.2 & 72.7\% $\pm$ 4.9 & \best{82.5\% $\pm$ 4.5} & \best{68.5\% $\pm$ 5.2} & 51.9\% $\pm$ 5.5 \\
& MIPROv2  & \best{72.7\% $\pm$ 5.2} & 69.8\% $\pm$ 5.2 & \best{73.4\% $\pm$ 5.2} & 81.5\% $\pm$ 4.5 & 65.3\% $\pm$ 5.5 & 50.3\% $\pm$ 5.5 \\
\bottomrule
\end{tabular}
\caption{MedHELM leaderboard (medical) across six language models and five prompting methods. \textbf{Green} marks the highest performance among evaluated prompting methods. Entries are reported as mean $\pm$ 95\% bootstrap CI.}
\label{tab:medhelm_benchmarks}
\end{table*}  

\subsection{Generalizability beyond DSPy}
To study whether leaderboard sensitivity persists under non-DSPy structured prompting strategies, we evaluate GEPA~\citep{agrawal2025gepa}, a reflective prompt optimizer. We use GEPA's standalone implementation rather than the DSPy instantiation to ensure that our findings are not an artifact of DSPy. On MMLU-Pro, across most models, GEPA demonstrates improvements over baseline (\textit{Claude 3.7}: 79.0\%, \textit{Gemini 2.0}: 69.5\%, \textit{GPT 4o}: 73.2\%, \textit{o3 Mini}: 77.6\%, \textit{Llama 3.3}: 63.4\%, \textit{Qwen3 4B}: 55.1\%), reinforcing that the choice of optimizer is less important than the conclusion that leaderboards can shift under alternative prompts.

\subsection{Generalizability Beyond Multiple-Choice}
To test whether our findings extend beyond multiple-choice tasks, we evaluate \textit{Qwen3-4B} on IFEval~\citep{zhou2023instruction}, an instruction-following benchmark with deterministic rule-based evaluation. Accuracy improves from 65.4\% with Zero-Shot Predict to 75.3\% with Zero-Shot CoT, 78.4\% with GEPA, and 78.8\% with MIPROv2. This confirms that CoT provides the strongest cost-performance tradeoff, while optimization can offer meaningful additional gains for smaller open-source models. Notably, \textit{Qwen3-4B} with MIPROv2 slightly exceeds \textit{DeepSeek-R1 Thinking} (78.4\%) and approaches frontier LMs on the HELM public leaderboard, including \textit{Claude 4.5 Haiku} (80.1\%), \textit{Qwen3-Next 80B A3B Thinking} (81.0\%), \textit{Llama 3.1 405B Instruct Turbo} (81.1\%), \textit{Qwen3 235B A22B Thinking} (81.6\%), and \textit{GPT-4o} (81.7\%), illustrating that the same small model can appear more competitive to frontier models when invoked with a structured prompt configuration.

\subsection{Controlled Prompt Variation Analysis}
\label{subsec:prompt_variation_analysis}

To isolate the effect of CoT, we report mean $\pm$ standard deviation across groups and compare matched non-CoT variants (HELM baseline, BFRS without CoT, and MIPROv2 without CoT) against CoT-based variants (Zero-Shot CoT, BFRS, and MIPROv2), holding instructions and demonstrations the same. On MedCalc-Bench, we run this analysis for \textit{GPT 4o} and \textit{Qwen3 4B}. For \textit{Qwen3 4B}, non-CoT variants achieve 8.7\% $\pm$ 3.5\% accuracy, compared with 21.7\% $\pm$ 1.0\% for CoT-based variants. For \textit{GPT 4o}, non-CoT variants achieve 19.9\% $\pm$ 1.0\%, compared with 26.8\% $\pm$ 0.2\% under CoT, suggesting that once the CoT interface is introduced, performance is higher and more stable.

\begin{figure}[t]
  \centering
  \includegraphics[width=0.41\textwidth]{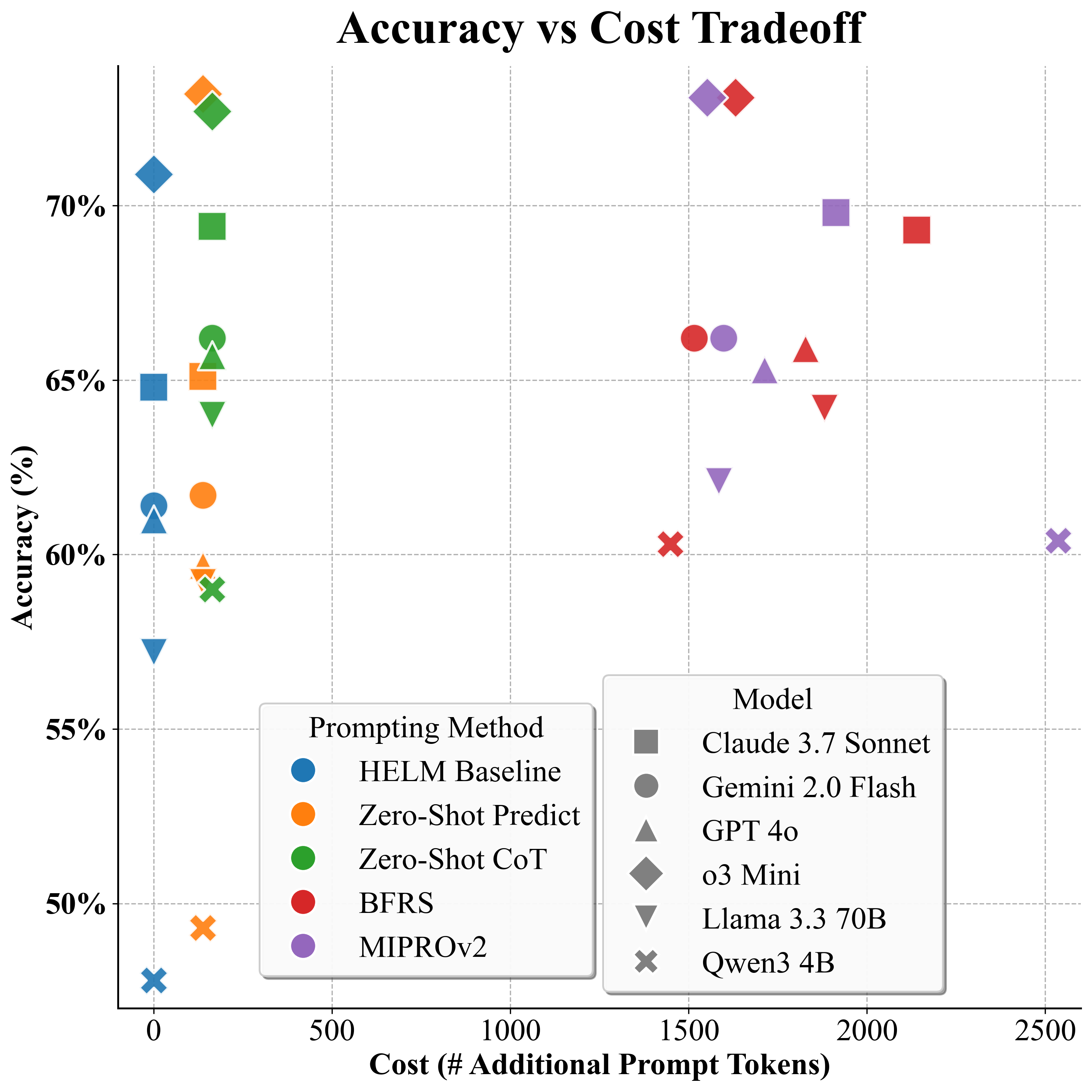}
  \footnotesize
  \caption{Accuracy vs cost tradeoff across prompting methods. Each point represents a model-prompt pair, with x-axis showing additional prompt tokens (relative to HELM baseline) and y-axis showing macro-averaged accuracy across benchmarks. Zero-Shot CoT achieves the most cost-effective improvements.}
  \label{fig:cost_analysis}
\end{figure}

\begin{figure*}[t]
  \centering
  \includegraphics[width=1.0\textwidth]{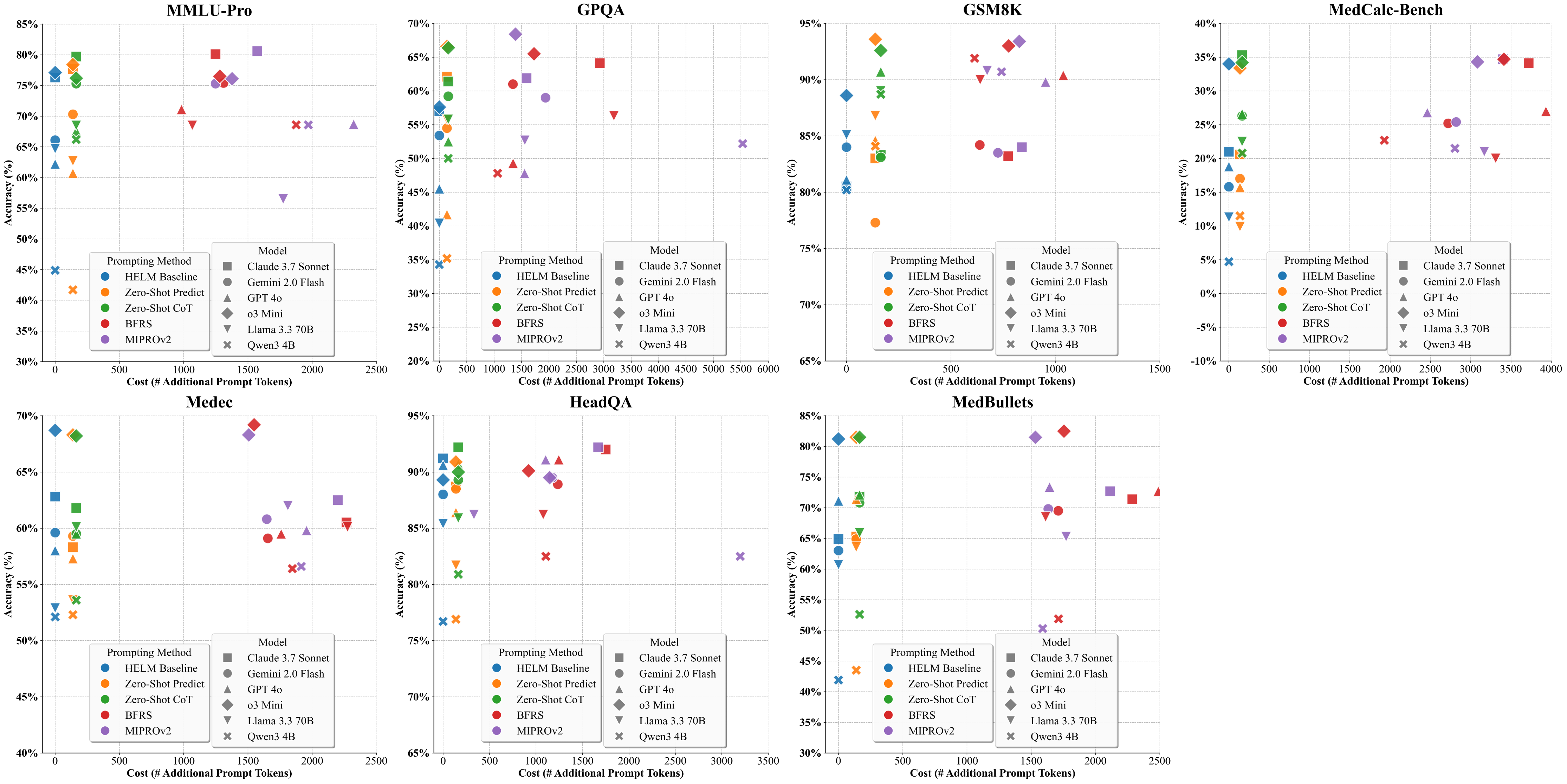}
  \footnotesize
  \caption{Accuracy vs cost tradeoff across prompting methods and benchmarks. Each point represents a model-prompt pair, with x-axis showing additional prompt tokens (relative to HELM baseline) and y-axis showing accuracy. Zero-Shot CoT achieves the most cost-effective improvements.}
  \label{fig:cost_analysis_granular}
\end{figure*}

\subsection{Computational Cost Analysis}

We evaluate the computational cost of prompting methods through inference-time token usage. We focus on recurring inference-time cost (relevant to leaderboard reporting and deployment), and treat prompt optimization as an amortized one-time setup cost because even the most expensive optimization run cost below \$10 (small relative to the cost of evaluating full benchmark suites). In our setup, the input (e.g., question, patient note) is identical across prompting methods, and outputs are capped at $<$200 tokens. The differences in inference cost arise almost entirely from the \emph{prompt prefix} (the instructions and demonstrations prepended to the input). Hence, we quantify the number of \emph{additional prompt tokens} relative to the HELM baseline instruction, capturing each prompting method's overhead. DSPy introduces a lightweight structured prompt template across all methods, resulting in 138 additional tokens for Zero-Shot Predict and 164 tokens for Zero-Shot CoT, which further includes a brief reasoning header. In contrast, BFRS and MIPROv2 insert task-specific demonstrations, producing much larger prompts: averaged across LMs and benchmarks, BFRS adds 1,741 tokens per query and MIPROv2 adds 1,816. Figure~\ref{fig:cost_analysis} shows the resulting tradeoff aggregated across benchmarks. Few-shot optimizers reach high performance but require the largest token budgets. Zero-Shot CoT captures most of these gains while using minimal additional prompt tokens, making Zero-Shot CoT the most cost-effective structured prompting method in our study. Figure~\ref{fig:cost_analysis_granular} shows the resulting tradeoff at a more granular level (across each benchmark, prompting method, and model), reinforcing the cost-performance tradeoff of Zero-Shot CoT.

\section{Related Work}

\textbf{Holistic benchmarking.} The General Language Understanding Evaluation (GLUE)~\citep{wang2018glue} was one of the first multi-task evaluation frameworks, aggregating nine language understanding tasks. Benchmarks of increasing scale followed: (i) Measuring Massive Multitask Language Understanding (MMLU)~\citep{hendrycks2021measuringmassivemultitasklanguage}, including 57 tasks spanning STEM, humanities, social sciences, and (ii) Beyond the Imitation Game (BIG-Bench)~\citep{srivastava2023imitationgamequantifyingextrapolating}, with 204 diverse tasks. The HELM framework is an established standard, designed for transparent and reproducible evaluation of LMs~\citep{liang2022holistic}. However, these benchmarks are typically evaluated using a single static prompt configuration.~\citet{liang2022holistic} note they opt for simple, generic prompts to orient development "towards generic language interfaces" that do not require "model-specific incantations". This reliance on a static prompt configuration, however, can underestimate LM capabilities~\citep{srivastava2023imitationgamequantifyingextrapolating, suzgun2022challengingbigbenchtaskschainofthought}.~\citet{mizrahi2024state} further demonstrate score and ranking instability across semantically equivalent instruction paraphrases and propose metrics. We study a distinct axis within HELM (structured prompting) and compare no-CoT, CoT, selected demonstrations, and optimized instructions alongside their cost-performance tradeoffs.

\textbf{Prompting methods.} The discovery of in-context learning~\citep{brown2020languagemodels}, where models learn from n-shot demonstrations, and the breakthrough of chain-of-thought (CoT) prompting~\citep{wei2022chainofthought} established the important role of prompt design in model performance. Complex, manually-composed strategies like Medprompt~\citep{nori2023medprompt}, which combine few-shot selection, CoT, and ensembling, demonstrate that LM performance can be substantially higher than under a single static prompt configuration. Because manual prompt engineering is impractical to apply systematically, researchers often frame prompt design as a formal "optimization problem", leading to the field of APO. Early APO methods include generation-and-selection, such as Automatic Prompt Engineer (APE)~\citep{zhou2023largelanguagemodelshumanlevel}, which uses an LM to propose candidate instructions and a separate scoring function to select the best one. Subsequent systems expanded this search paradigm~\citep{wang2023promptagent, yang2023large, singla2024dynamic}. These methods often outperform zero-shot or manually engineered prompts across tasks. In the LM-as-Optimizer paradigm, an LM is instructed to iteratively refine prompts by showing it a trajectory of previously evaluated candidates. Other approaches have employed evolutionary search, like Promptbreeder~\citep{fernando2024promptbreeder}, which treats prompts as "genes" and evolves a population of instructions over generations. The DSPy framework~\citep{khattab2023dspy} generalizes these methods, providing a programming model that compiles declarative, multi-stage pipelines.

\section{Conclusion}
We systematically study how prompt choice impacts reported performance within an established benchmarking framework. Our results show that prompt choice can materially alter benchmark conclusions, shifting relative LM ordering and altering the gaps between models. Sensitivity is heterogeneous: reasoning LMs show marginal gains, whereas some benchmarks for non-reasoning LMs benefit more. A striking example is \textit{Qwen3 4B}, which demonstrates the largest improvement moving from baseline to the best-performing prompt, matching several frontier baselines and reiterating that static prompt configurations may not provide a complete picture for deployment decisions. Another key finding is that gains are largely agnostic to the particular structured prompting method once CoT is introduced. Public leaderboards should report performance under multiple prompts, enabling practitioners to assess how strongly conclusions depend on prompt choice. Broadly, evaluating a model under multiple prompts can provide a more holistic picture than static prompt configurations.

\section{Limitations}
We evaluate six LMs spanning frontier and open-source systems; broader coverage across model families and sizes is needed to establish generality. Our benchmarks primarily involve multiple-choice and short reasoning tasks, where evaluation is deterministic and reproducible and therefore well suited to isolating the effect of prompting on leaderboard outcomes. Open-ended generation settings typically rely on learned or LM-based evaluators, which introduce additional evaluator sensitivity and optimization-evaluation mismatch; prior work also suggests that CoT gains are most reliable on math and symbolic reasoning, with more mixed effects in long-form generation~\citep{sprague2024cot}. We study a subset of standard prompting methods rather than exhaustively searching the prompt space. Our controlled prompt-variation analysis likewise isolates one dimension of prompting and should be viewed as targeted evidence rather than a comprehensive perturbation study. We use single fixed-seed, zero-temperature optimization and evaluation runs to preserve a reproducible HELM-style protocol. Bootstrap CIs quantify uncertainty over evaluation examples, not optimizer-search variance across seeds or splits; multi-seed and multi-split optimization remain future work.

\clearpage

\section*{Data Availability}
The public datasets used in this study are: MMLU-Pro~\citep{wang2024mmlu}, GPQA~\citep{rein2024gpqa}, GSM8K~\citep{cobbe2021gsm8k}, MedCalc-Bench~\citep{khandekar2024medcalc}, Medec~\citep{abacha2024medec}, HeadQA~\citep{vilares2019head}, MedBullets~\citep{medbullets}, and IFEval~\citep{zhou2023instruction}. Further distribution is subject to the sharing agreements stipulated by the creators.

\section*{Acknowledgments}
AA is supported by NIH grant R01 HL167974 and ARPA-H contract AY2AX000045. NHS acknowledges support from the Debrah and Mark Leslie endowment for AI in Healthcare, and salary support from Stanford Healthcare. ASC receives research support from NIH grants R01 HL167974, R01HL169345, R01 AR077604, R01 EB002524, R01 AR079431, P41 EB027060; ARPA-H contracts AY2AX000045 and 1AYSAX0000024-01; and NIH contracts 75N92020C00008 and 75N92020C00021. Unrelated to this work, ASC receives research support from GE Healthcare, Philips, Microsoft, Amazon, Google, NVIDIA, Stability; has provided consulting services to Patient Square Capital, Chondrometrics GmbH, and Elucid Bioimaging; is co-founder of Cognita; has equity interest in Cognita, Subtle Medical, LVIS Corp, Brain Key, and Radiology Partners. This research was, in part, funded by the Advanced Research Projects Agency for Health (ARPA-H). The views and conclusions contained in this document are those of the authors and should not be interpreted as representing the official policies, either expressed or implied, of the United States Government.

\section*{Competing Interest}
No competing interests to declare.

\section*{Ethical Considerations}
Our findings may create incentives to further optimize prompts for benchmark performance, which could encourage leaderboard gaming if evaluation protocols do not standardize prompt families, report prompt-search budgets, or summarize performance across multiple prompt settings. In addition, some medical prompts used in this work employ high-stakes framing for benchmarking consistency; these prompts are not intended for clinical use.

\clearpage
\bibliography{custom}
\clearpage

\end{document}